\documentclass[english]{lni}

\usepackage{graphicx}
\usepackage{fancyhdr}
\usepackage{changepage} 
\usepackage{listings} 
\usepackage{amsmath}
\usepackage{float}
\usepackage[figurename=Fig., tablename=Tab.]{caption}
\usepackage[utf8]{inputenc}  
\usepackage{newunicodechar}
\usepackage{refcount}
\usepackage{booktabs}
\usepackage{siunitx}

\newunicodechar{≈}{\ensuremath{\approx}}

\renewcommand{\headrulewidth}{0.4pt} 
\author{Alfonso Pedro Ridao\footnote{Department of Applied Mathematics and Computer Science, Technical University of Denmark (DTU), s243942@student.dtu.dk}}
\title{A Quantitative Method for Shoulder Presentation Evaluation in Biometric Identity Documents}

\begin{document}

\maketitle

\setcounter{footnote}{1} 
\pagestyle{fancy} 
\fancyhead{} 

\fancyhead[LE]{\small \thepage \hspace{25pt} Alfonso Pedro Ridao}
\fancyhead[RO]{\small A Quantitative Method for Shoulder Presentation Evaluation \hspace{25pt} \thepage}
\fancyfoot{} 
\renewcommand{\headrulewidth}{0.4pt} 

\begin{abstract}
International standards for biometric identity documents mandate strict compliance with pose requirements, including the square presentation of a subject's shoulders. However, the literature on automated quality assessment offers few quantitative methods for evaluating this specific attribute. This paper proposes a Shoulder Presentation Evaluation (SPE) algorithm to address this gap. The method quantifies shoulder yaw and roll using only the 3D coordinates of two shoulder landmarks provided by common pose estimation frameworks. The algorithm was evaluated on a dataset of 121 portrait images. The resulting SPE scores demonstrated a strong Pearson correlation (r ≈ 0.80) with human-assigned labels. An analysis of the metric's filtering performance, using an adapted Error-versus-Discard methodology, confirmed its utility in identifying non-compliant samples. The proposed algorithm is a viable lightweight tool for automated compliance checking in enrolment systems.
\end{abstract}
\begin{keywords}
Shoulder Presentation, Biometric Quality, ISO/IEC 29794-5, Pose Estimation.
\end{keywords}


\section{Introduction}

For biometric identity documents like passports and ID cards, the use of high-quality frontal face photographs is a foundational requirement. To guarantee global consistency, international standards impose firm constraints on posture, with the latest ISO/IEC 29794-5 and ICAO Portrait Quality guidance being the key specifications~\cite{iso29794-5-2025, icao2007}. These standards mandate a frontal head orientation within a narrow tolerance window (e.g., ±5° yaw/pitch, ±8° roll)~\cite{iso29794-5-2025}. Critically, they also require that the shoulders be presented squarely to the camera and parallel with the image plane~\cite{iso29794-5-2025}. This strictness is justified by large-scale studies from institutions like the National Institute of Standards and Technology (NIST), which consistently identify pose as one of the most critical quality factors impacting face recognition performance~\cite{grother2014frvt}.

Control of biometric sample quality thus hinges on ensuring adherence to these pose constraints. Previously, this task fell to human screeners, a process that was labour-intensive, slow, and subjective at scale~\cite{ferrara2008evaluating}. Today, modern enrolment kiosks and e-gates integrate automatic quality modules to reject non-compliant images on the spot. However, even state-of-the-art toolkits, such as the FaceQvec vector-based quality model, primarily deliver scores for head pose but do not yet offer a distinct, interpretable metric for the presentation of the shoulders~\cite{hernandez2022faceqvec}.

This is a notable gap, as shoulder squareness is not always coupled with head yaw. A person might turn their head while keeping their torso frontal, or conversely, twist their shoulders even when their face remains centered. A truly robust compliance system must therefore verify two conditions: that head pose angles remain within tolerance, and that the shoulder line is parallel to the image plane. The new ISO/IEC 29794-5:2025 standard acknowledges this by specifying ``shoulder presentation'' as a standalone quality measure (measure \#32 in Table 3)~\cite{iso29794-5-2025}.

To address this gap, the Shoulder Presentation Evaluation (SPE) algorithm is proposed as a method to quantify how orthogonally a subject’s shoulders are oriented to the camera’s image plane. The algorithm operates on the 3D world coordinates for the left- and right-shoulder landmarks, which are supplied in real time by Google’s lightweight MediaPipe Pose estimator~\cite{bazarevsky2020blazepose}. From just these two 3D points, a pair of interpretable posture metrics are derived:
\begin{enumerate}
    \item A horizontal-alignment (yaw) score, which is calculated from the relative depth (z-coordinate) difference between the shoulders to reflect left--right rotation.
    \item A tilt (roll) score, which is calculated from the vertical displacement of the shoulders to capture whether one is raised higher than the other.
\end{enumerate}

Following their computation, each metric is normalised to the unit interval [0, 1], where 1 signifies perfect conformance and 0 represents a completely non-compliant posture. An overall ``shoulder squareness'' score is then produced from their geometric mean, providing a single value suitable for an operator or an automated rejection system. Since the analysis depends on only two landmarks, SPE achieves real-time performance of over 100 fps (see Section 3.3 for benchmarks), making it practical for providing instant user feedback in self-service kiosks and for implementing the new ISO/IEC 29794-5 quality check~\cite{iso29794-5-2025, bazarevsky2020blazepose}.

\section{Related Work}

International standards have long recognized the importance of pose in biometric photography. The ISO/IEC 19794-5:2005 standard first codified the need for a frontal face and explicitly prohibited ``portrait style'' over-the-shoulder poses~\cite{iso19794-5-2005}. The superseding ISO/IEC 39794-5:2019 standard refined this with more detailed geometric requirements for both head and shoulder orientation to ensure a fully frontal presentation~\cite{iso39794-5-2019}. A nearly identical pose envelope is specified in ICAO Doc 9303, which limits head rotation to a few degrees~\cite{icao2007}. The forthcoming ISO/IEC 29794-5:2025 quality standard reinforces this by introducing a dedicated ``shoulder presentation'' metric~\cite{iso29794-5-2025}. Early evaluation frameworks, such as that of Ferrara et al. (2008), highlighted the practical challenges of enforcement; while their work acknowledged the standard's requirements, their test battery lacked a quantitative metric for shoulder alignment, focusing instead on head-pose angles and other photographic properties~\cite{ferrara2008evaluating}.

Accurate quality assessment is therefore dependent on robust pose estimation. Early work in this field often derived pose angles from the geometry of sparse 2D facial landmarks. In contrast, modern systems like Google's Attention Mesh predict a dense 3D face mesh in real-time for highly precise pose extraction~\cite{grishchenko2020attentionmesh}. These estimates are an integral part of modern Face-Image Quality Assessment (FIQA) systems. As a comprehensive survey by Schlett et al. confirms, head pose is one of the most influential factors for face recognition error rates~\cite{schlett2022survey}. The FaceQvec toolkit, for instance, incorporates a specific test for head-pose angles as part of its compliance checks~\cite{hernandez2022faceqvec}.

For all the progress in head-pose estimation, the literature offers almost no quantitative treatment of shoulder orientation as a distinct quality dimension. The prevailing approach in many quality pipelines is to assume that the torso follows the head; if head-pose angles are within tolerance, the portrait is accepted as frontal. This assumption, however, can be unreliable in user-controlled capture scenarios. To address this gap, full-body landmark detectors such as MediaPipe Pose~\cite{bazarevsky2020blazepose} and OpenPose~\cite{cao2021openpose} can provide the necessary 2D or 3D shoulder coordinates. While these toolkits are widely applied in other domains, a search of the literature revealed no prior work that converts their outputs into a continuous ``shoulder-squareness'' metric for biometric compliance. The Shoulder Presentation Evaluation (SPE) algorithm proposed in this paper therefore fills a demonstrable gap by augmenting existing FIQA models that penalize head angles but overlook the critical dimension of torso alignment.

The value of any new biometric quality score is determined by its ability to predict recognition failures. The standard methodology is the Error-versus-Discard Characteristic (EDC) curve, introduced by Grother and Tabassi~\cite{grother2007performance} and now recommended in the ISO/IEC 19795 series of standards~\cite{iso19795-1-2021}. An EDC curve plots a recognition error rate, such as the False Non-Match Rate (FNMR), against the fraction of lowest-quality samples being discarded. This type of analysis is central to modern biometrics. For example, it underpins the NFIQ 2.0 standard for fingerprints~\cite{tabassi2021nfiq2} and is a core component of the face-quality evaluation framework developed by Schlett et al.~\cite{schlett2024considerations}. An effective quality metric will show a steep decline in error after removing only a small fraction of poor-quality samples. This paper adapts the EDC methodology to evaluate how effectively the proposed SPE score ranks image compliance against human-adjudicated labels.

\section{Methodology}

The Shoulder-Presentation Evaluation (SPE) routine quantifies how square a traveller’s torso is to the camera by combining two complementary metrics. The algorithm operates on the 3D coordinates of the left (L) and right (R) shoulder landmarks (indices 11 and 12) provided by Google’s lightweight MediaPipe Pose runtime~\cite{bazarevsky2020blazepose} to compute:
\begin{itemize}
    \item \textbf{Horizontal-alignment score ($s_{\text{yaw}}$)} – The cosine of the estimated yaw angle, derived from the relative depth ($z$) and horizontal separation ($x$) of the shoulders. It reflects left-right rotation and is penalized by landmark visibility.
    \item \textbf{Shoulder-tilt score ($s_{\text{roll}}$)} – A linear mapping of the absolute roll angle between the shoulders, obtained from the arctangent of the vertical and horizontal separations. It captures how level the shoulders are (roll).
\end{itemize}

After normalising each term to the unit interval [0, 1], a single ``shoulder-squareness'' quality value $q$ is produced as their geometric mean:
\begin{equation}
    q = \sqrt{s_{\text{yaw}} \cdot s_{\text{roll}}}
\end{equation}

While this combined score is suitable for automated rejection, the individual components are retained for providing clearer, more specific user feedback (e.g., distinguishing a yaw issue from a roll issue). This work's analysis therefore considers both the independent components for diagnostic clarity and the final combined score for holistic performance evaluation. The core logic of the algorithm is summarized in the pseudocode below, with the full implementation detailed in the supplementary material\footnote{\label{note:supplementary}The complete project, including source code, datasets, and a reproducible analysis notebook, is submitted with this paper in a digital archive and is also available in the project's public repository at: https://github.com/fonCki/shoulder-presentation-eval/tree/main/supplementary-material}. This approach remains fully compatible with ISO/IEC 29794-5 measure 32 for ``shoulder presentation''~\cite{iso29794-5-2025}. Its reliance on only two landmarks enables real-time performance ($\approx$ 0.2 ms per frame\footnote{An open-source implementation of the SPE algorithm is publicly available at https://github.com/fonCki/shoulder-presentation-eval. A live web demonstration, which allows for real-time testing with a camera or uploaded images, can be accessed at https://shoulder-presentation-eval.ridao.ar.}, making it practical for enrolment kiosks.

\vspace{1em}

\begin{center}
\textbf{Pseudocode for the SPE Algorithm}
\end{center}
\begin{verbatim}
horizontalScore(L, R):
    dx <- L.x - R.x                   // Horizontal gap
    dz <- L.z - R.z                   // Depth gap
    if ||(dx,dz)|| < epsilon: return 0 

    geom  <- |dx| / sqrt(dx^2 + dz^2) // Cosine of the yaw angle
    vis   <- 1 - |v_L - v_R|      // Visibility symmetry penalty
    return geom * vis

shoulderTiltScore(L, R):
    dx <- L.x - R.x
    dy <- L.y - R.y          // Vertical separation of shoulders
    if max(|dx|,|dy|) < epsilon: return 0

    angle <- arctan(|dy| / |dx|)     // Roll angle in radians
    return 1 - angle / (pi/2)        // Map angle to score [0,1]
\end{verbatim}
\vspace{1em}

\pagebreak

\subsection{Horizontal-alignment score (yaw)}

Let the normalised landmark coordinates returned by MediaPipe Pose be $L = (x_L, y_L, z_L)$ and $R = (x_R, y_R, z_R)$. A purely frontal torso is characterised by (i) equal depth, $z_L \approx z_R$, and (ii) a wide horizontal separation, $|x_L - x_R|$. Conversely, as the subject rotates about the vertical axis, the depth gap $|\Delta z| = |z_L - z_R|$ grows while the apparent horizontal distance $|\Delta x| = |x_L - x_R|$ shrinks. These two cues form a right-angled triangle in the camera’s horizontal plane, from which the yaw angle is obtained as:
\begin{equation}
    \theta = \arctan( |\Delta z| / |\Delta x| ) \quad (0^\circ \leq \theta \leq 90^\circ)
\end{equation}
Its cosine,
\begin{equation}
    \cos \theta = \frac{|\Delta x|}{\sqrt{(\Delta x)^2 + (\Delta z)^2}},
\end{equation}
is therefore a geometry term that scores 1 for a perfectly square torso ($\theta \approx 0^\circ$) and 0 for a profile view ($\theta = 90^\circ$). Because landmark reliability deteriorates when one shoulder is occluded, this term is weighted by a visibility-symmetry factor $v_{\text{sym}} = 1 - |v_L - v_R|$, where $v_L$ and $v_R$ are the per-landmark confidence scores that accompany the Pose output~\cite{bazarevsky2020blazepose}. The final metric is thus:
\begin{equation}
    \text{horizontalScore} = \cos \theta \times v_{\text{sym}},
\end{equation}
which remains in [0, 1]. High values are produced only when the shoulders are widely separated in the image, lie at similar depth, and are detected with comparable confidence. This aligns with the yaw-quality criteria adopted in the NIST Face Recognition Vendor Test (FRVT)~\cite{grother2014frvt}. The intermediate quantity $\theta$ is retained for debugging purposes.

\subsection{Shoulder-tilt score (roll)}

Even when the torso is facing the camera, a subject may lean sideways so that one shoulder is higher than the other. The ISO/IEC face-image quality requirements expect the shoulder line to be level, mirroring the ``no-roll'' requirement for the head~\cite{iso29794-5-2025}. To quantify this, the SPE algorithm projects the shoulder baseline onto the image plane. The vertical offset is $\Delta y = y_L - y_R$; if $\Delta y = 0$ the shoulders are perfectly horizontal. Using the horizontal span $|\Delta x|$ as the baseline, the roll angle is:
\begin{equation}
    \phi = \arctan( |\Delta y| / |\Delta x| ) \quad (0^\circ \leq \phi \leq 90^\circ),
\end{equation}
so that $\phi = 0^\circ$ when the shoulders are level and $\phi = 90^\circ$ when they form a vertical line. This is mapped to a unit-interval score:
\begin{equation}
    \text{shoulderTiltScore} = 1 - (\phi / 90^\circ) = 1 - \frac{2 \cdot \arctan(|\Delta y|/|\Delta x|)}{\pi},
\end{equation}
yielding 1 for a perfectly horizontal shoulder line and tending toward 0 as the tilt approaches an extreme roll. In practical portrait capture, most compliant images achieve a shoulderTiltScore $\approx$ 1; values below 0.8 may indicate a slumped posture or a tilted camera.

\subsection{Implementation and Ground Truth}

The prototype was written in TypeScript as a React single-page application. Each incoming video frame is processed by the in-browser MediaPipe Pose runtime, which produces a 33-landmark skeleton in a few milliseconds\footnote{Performance benchmarked on an Apple M1 Pro CPU running the algorithm in a standard web browser} using its WebAssembly/GPU backend~\cite{bazarevsky2020blazepose}. Only the two shoulder points are retained, after which the horizontal- and tilt-scores are computed with the closed-form equations.

Ground-truth labels were created by visually inspecting every portrait and assigning a subjective score in [0, 1]: 1 for a perfectly frontal torso, and $\approx$0.1–0.2 for the most rotated samples. A seed set of 15 images with known poses was used to calibrate the score mapping and validate the model's scale. The exercise confirmed that a $\approx$45° shoulder rotation yields horizontalScore $\approx$ 0.5 and led to the introduction of the $v_{\text{sym}}$ term, as early experiments with the raw $\cos \theta$ proved too lenient when one landmark was occluded. To handle cases of extreme rotation robustly, the algorithm defaults to a score of 0 if either shoulder landmark is undetected or falls below a confidence threshold (see the supplementary material\footnotemark[\getrefnumber{note:supplementary}]), ensuring non-compliant poses receive a failing grade.

A further advantage of SPE is that it makes no use of facial landmarks or a face detector. Consequently, it can return a meaningful quality value when the face is masked, partially occluded, or turned away, as long as the shoulders remain visible. This decoupling is consistent with the upcoming ISO/IEC 29794-5 draft, which treats ``shoulder presentation'' as a quality measure independent of head-pose compliance~\cite{iso29794-5-2025}.

\begin{figure*}[h]
    \centering
    \includegraphics[width=\textwidth]{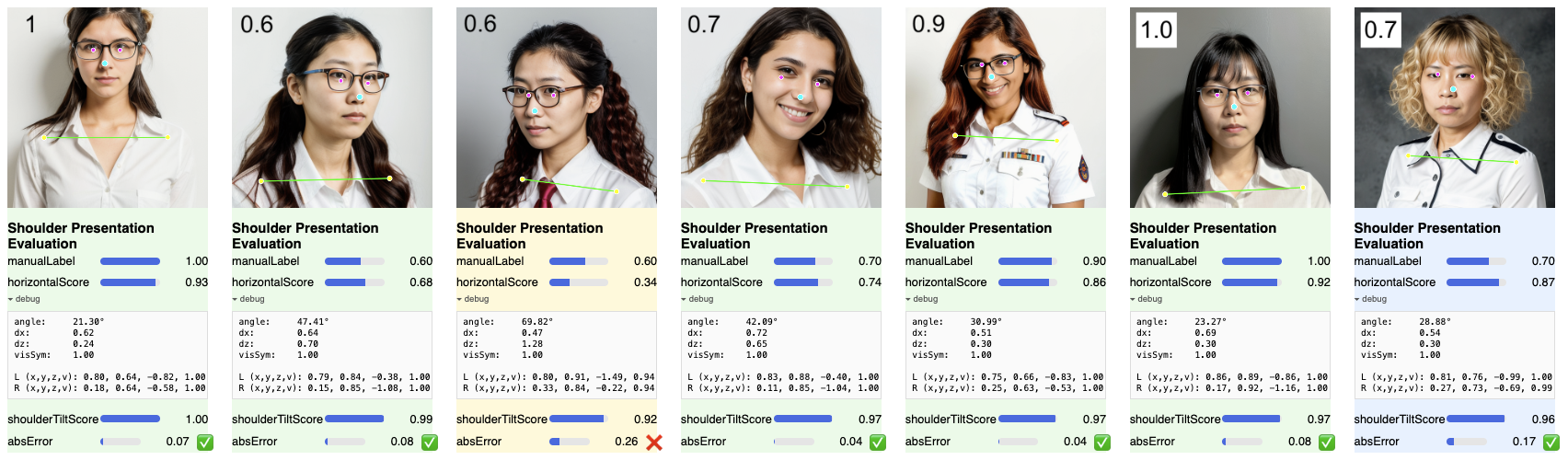}
    \caption{Example results of the Shoulder-Presentation Evaluation (SPE) algorithm. Each of the seven panels shows a test image with MediaPipe Pose landmarks overlaid. The evaluation panel below each image compares the algorithm's computed horizontalScore and shoulderTiltScore against the human-annotated manualLabel (ground truth). Debug information, including the calculated yaw angle and the absError between the horizontalScore and manualLabel, illustrates the model's internal calculations. The examples range from fully compliant (left, score = 1.0) to moderately rotated torsos, demonstrating a strong correlation between the algorithm's scores and perceptual judgments.}
    \label{fig:results_example}
\end{figure*}

\section{Experimental Evaluation}

\subsection{Dataset and Ground-Truth Labels}

The evaluation set contains 122 portrait photographs\footnotemark[\getrefnumber{note:supplementary}]. One image was excluded from the quantitative analysis due to an interpretation error, resulting in a final test set of 121 samples. A core set of 102 images was sourced from the ``TONO-mixed'' collection of the TONO synthetic dataset~\cite{borghi2025tono}. The TONO dataset is specifically designed for evaluating ISO/ICAO compliance, providing images with a range of controlled non-conformities. The ``-mixed'' subset was chosen as it represents a more challenging and realistic scenario, containing images that may violate one or more compliance requirements. To further augment the diversity of shoulder poses, particularly for near-profile views, an additional 20 portraits were captured in-house from two volunteer subjects rotating through 360 degrees.

From the 104 images available in the ``TONO-mixed'' collection, two files were discarded due to data corruption. Crucially, no further filtering or pre-selection was performed on the remaining images. This approach ensures an unbiased evaluation of the algorithm's performance across the full spectrum of frontal and near-frontal poses provided by the dataset authors.

To establish a ground truth, each of the 121 photographs with a valid label was assigned a subjective shoulder-yaw score, $(y)$. Given that shoulder tilt was minimal across the dataset (as discussed in Section~4.2), the author, acting as a single, consistent rater, focused the judgment on the degree of left-right shoulder rotation (yaw). A discrete label was assigned from the set $\{0.0, 0.1, \dots, 0.9, 1.0\}$, where a score of 1.0 indicates a perfectly frontal shoulder presentation (no yaw) and 0.0 represents a full profile view. While this single-rater process provides a consistent baseline for this study, it is acknowledged that a more robust ground truth could be established in future work through a larger-scale study involving multiple raters or a crowd-sourced evaluation. The resulting labels serve as the reference values against which the Shoulder-Presentation Evaluation (SPE) scores are compared.

\begin{table}[btp]
    \centering
    \caption{Performance metrics of the SPE algorithm on 121 test images.}
    \label{tab:perf_metrics}
    \begin{tabular}{l l}
        \toprule 
        \textbf{Metric} & \textbf{Value} \\
        \midrule 
        \multicolumn{2}{l}{\textit{Regression Analysis}} \\
        \quad Pearson Correlation (r) & 0.801 \\
        \quad Mean Absolute Error (MAE) & 0.099 \\
        \quad Root Mean Squared Error (RMSE) & 0.162 \\
        \midrule 
        \multicolumn{2}{l}{\textit{Classification Analysis ($\tau=0.8$)}} \\
        \quad True Positives (TP) & 66 \\
        \quad True Negatives (TN) & 34 \\
        \quad False Positives (FP) & 5 \\
        \quad False Negatives (FN) & 16 \\
        \quad False Positive Rate (FPR) & 12.82\% \\
        \quad False Negative Rate (FNR) & 19.51\% \\
        \bottomrule 
    \end{tabular}
\end{table}

As shown in Table~\ref{tab:perf_metrics}, the results indicate a strong positive linear relationship between the algorithm's score and the manual labels, confirmed by a Pearson correlation ($r \approx 0.801$) and a low Mean Absolute Error (MAE $\approx 0.099$).

\begin{figure}[h]
    \centering
    \includegraphics[width=0.7\columnwidth]{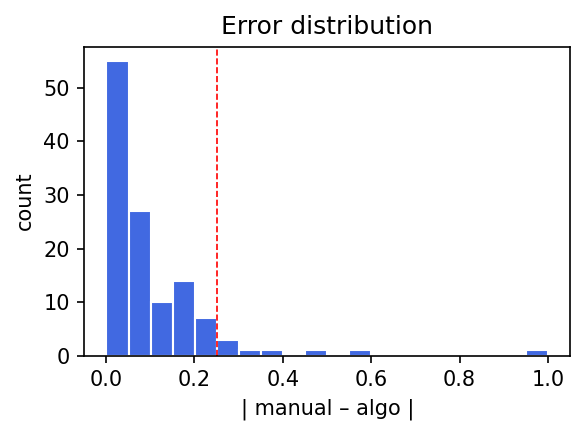}
    \caption{Distribution of absolute errors ($|$manual – algorithm score$|$) across the test set. The histogram shows that in the majority of cases, the error is very low. The red dashed line highlights an error threshold of 0.2.}
    \label{fig:error_dist}
\end{figure}

In the binary classification task (using a compliance threshold of $\tau=0.8$), the algorithm demonstrates a practical trade-off. The False Negative Rate (FNR) of \SI{19.5}{\percent} is higher than the False Positive Rate (FPR) of \SI{12.8}{\percent}, suggesting a conservative model that prefers to incorrectly reject a compliant image over accepting a non-compliant one. The error histogram in Figure~\ref{fig:error_dist} shows that deviations are minimal for most samples, though 13 images had an absolute error greater than 0.2. One notable outlier (error of 1.0) corresponded to an image with an extreme close-up crop where the shoulders were not visible. This caused the landmark detector to fail, producing a \texttt{horizontalScore} of 0.0, while the human rater assigned a score of 1.0. This case underscores the algorithm's dependency on the underlying landmarking quality.

\begin{figure}[h]
    \centering
    \includegraphics[width=0.6\columnwidth]{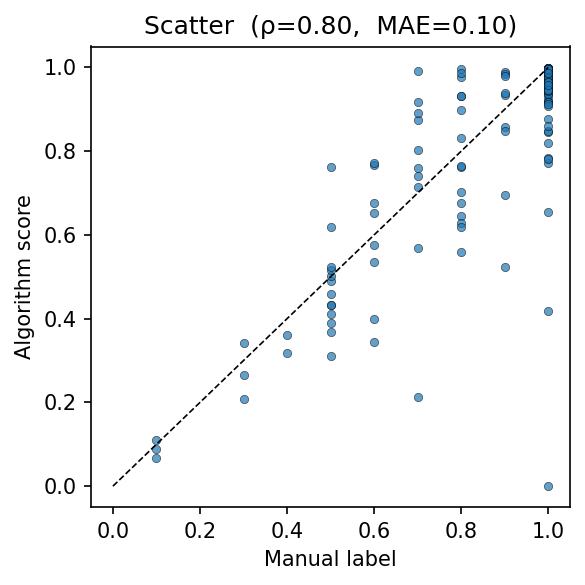}
    \caption{Scatter plot of the SPE algorithm’s output score vs. the manual label for shoulder alignment on 121 test images. The dashed line is the identity line (perfect prediction). The Pearson correlation is $\approx$0.80 and mean absolute error is 0.10, indicating a strong agreement between the automatic metric and human judgement.}
    \label{fig:scatter}
\end{figure}

\subsection{Shoulder-Tilt Score}
\label{sec:tilt}
In addition to the horizontal-alignment metric, SPE outputs a \texttt{shoulderTiltScore} that captures the vertical symmetry of the shoulder points (i.e., roll of the torso). An analysis of this metric across the evaluation set revealed that approximately \SI{83}{\percent} of the portraits had a score exceeding 0.90, reflecting both the subjects' natural tendency to maintain a level posture and the TONO dataset's focus on yaw-based non-compliance.

Because subtle shoulder height differences are difficult for a human to grade reliably from a 2D image, and given the lack of significant roll variation in the dataset, no explicit ground-truth labels were collected for this dimension. Instead, the focus of the manual labeling and primary evaluation was on the \texttt{horizontalScore}.

However, the \texttt{shoulderTiltScore} remains a crucial component for detecting upstream landmarking failures. One test image, where the subject's shoulders were occluded, caused the MediaPipe API to return erroneous, vertically-aligned landmark coordinates. The SPE algorithm correctly interpreted this invalid geometry, producing a very low \texttt{shoulderTiltScore} of 0.51. This case, the only notable failure observed for the tilt metric, demonstrates its utility as a sanity check: a low tilt score can effectively flag samples where the underlying landmark detection is unreliable, complementing the primary yaw check as specified in ISO/IEC FDIS 29794-5~\cite{iso29794-5-2025}.

\subsection{Adapted EDC Analysis for Compliance Filtering}

A primary goal of a biometric quality metric is to predict a sample's utility. The standard tool for this is the Error-versus-Discard Characteristic (EDC) curve, which plots a recognition system's False Non-Match Rate (FNMR) against the fraction of low-quality samples discarded~\cite{grother2007performance}. While the standard EDC curve is tied to a specific recognition system, the methodology can be adapted for a recognizer-agnostic evaluation.

To evaluate the SPE algorithm's ability to filter non-compliant samples, we adapt the standard EDC methodology. We replace the traditional matcher-dependent False Non-Match Rate (FNMR) with a classification False Negative Rate (FNR) calculated against human-provided ground-truth labels. This novel approach allows us to measure how effectively the algorithm's continuous quality score (\texttt{horizontalScore}) can be used to discard images that a human would deem non-compliant, without relying on a specific biometric matcher. While recognizer-agnostic evaluation methods have been explored~\cite{schlett2024considerations}, our use of a classification FNR against human labels is a direct assessment of the algorithm's alignment with human perception of compliance.

To construct the curve, we defined "error" as the False Negative Rate of the surviving population of images at each discard step. Specifically:
\begin{enumerate}
    \item Images are sorted by the algorithm's \texttt{horizontalScore} in ascending order (worst quality first).
    \item Samples are progressively "discarded" from this sorted list.
    \item At each discard fraction (x-axis), we recalculate the FNR on the \textbf{remaining} samples. This FNR (y-axis) is the fraction of compliant images (manual label $\ge 0.8$) that are misclassified as non-compliant by the algorithm (algo score $< 0.8$).
\end{enumerate}

\begin{figure*}[h]
    \centering
    \includegraphics[width=\textwidth]{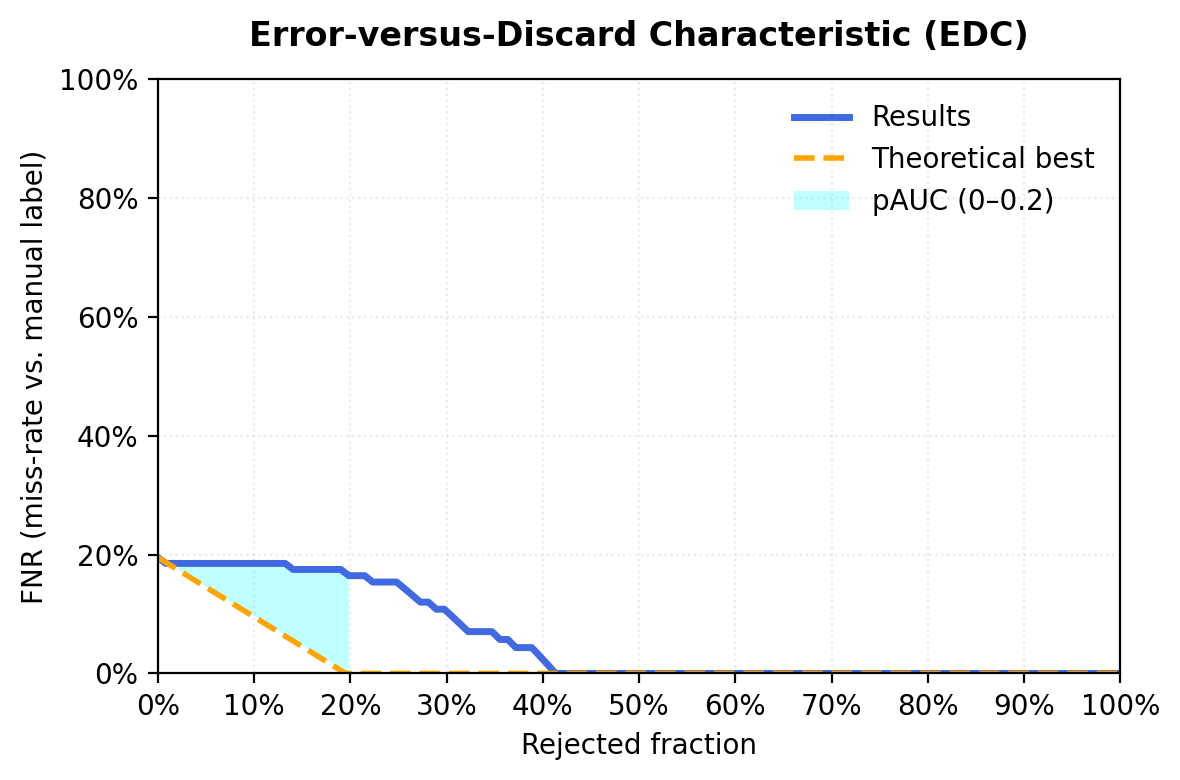}
    \caption{Adapted Error-versus-Discard Characteristic (EDC) for the shoulder-alignment quality score. The y-axis shows the \emph{classification} false-negative rate (FNR) relative to human labels, while the x-axis is the fraction of lowest-scored images discarded. This adapted EDC curve evaluates the metric's ability to filter non-compliant samples based on human judgment, rather than its effect on a specific biometric recognizer. The blue curve is the empirical SPE result; the orange dashed line is an oracle that would remove every misclassified image first. Discarding the worst 40\,\% of images lowers the FNR from 19.5\,\% to 4.3\,\%, illustrating the metric’s utility for compliance filtering.}
    \label{fig:edc}
\end{figure*}

As shown in Figure~\ref{fig:edc}, the initial FNR for the full dataset is \SI{19.5}{\percent}. The curve shows that discarding the lowest-scoring samples gradually improves this error rate. For example, discarding the \SI{20}{\percent} of images with the lowest \texttt{horizontalScore} reduces the FNR on the remaining population to \SI{17.5}{\percent}. To eliminate all false negatives, roughly \SI{40}{\percent} of the images must be discarded.

The orange dashed line represents a "perfect oracle" quality metric, which would discard all false negatives first, causing the error rate to drop to zero immediately. The gap between the empirical curve (blue) and the oracle curve indicates that the \texttt{horizontalScore} is a useful but imperfect predictor of human-labeled compliance. The relatively shallow initial slope of the curve suggests that many compliant images received a low score from the algorithm and are therefore only removed late in the discard process. While effective at identifying the most egregious non-compliant poses, the metric's ranking shows a moderate correlation with human judgment on borderline cases.

\subsection{Discussion}

These results confirm that the SPE algorithm is a viable tool for automatic pose compliance checking. The strong correlation with human judgements ($r \approx 0.80$) means it can serve as a proxy for a human inspector in flagging images where the subject’s pose is not acceptable. 

For a practical classification task with a strict threshold of $\tau=0.8$, the analysis revealed a False Negative Rate (FNR) of 19.5\%. A corresponding False-Positive Rate (FPR) of 12.8\,\% (Tab.~\ref{tab:perf_metrics}) completes the operating profile. This reflects a deliberate trade-off, preferring to inconvenience a small number of users by rejecting a compliant image (a false negative) over accepting a non-compliant one that could degrade system security (a false positive). This conservative operating point is further supported by the EDC analysis in Figure~\ref{fig:edc}. The curve demonstrates that the SPE score is a meaningful predictor of biometric utility, as progressively discarding the lowest-quality samples substantially reduces the proxy FNMR. For instance, removing the lowest‐scoring 20 \% of images cuts FNMR from 19.5\,\% to ≈17.5\,\%. This aligns with foundational findings in sample quality research~\cite{grother2014frvt, grother2007performance}. 

In practice—e.g. an online ID portal—SPE can give instant feedback and prompt retakes, cutting down the sub-par images that reach manual review.

\subsection{Limitations}

The primary limitations of this work are threefold. First, as a shoulder-only metric, SPE is not a complete posture compliance solution. It must be paired with an established head-pose check to achieve full ISO/IEC 29794-5 compliance, as it cannot detect the failure mode where the head is rotated while the shoulders remain square~\cite{grother2014frvt, hernandez2022faceqvec}. A practical implementation could address this via a multi-stage process, evaluating head pose before shoulder presentation to provide more targeted feedback.

Second, the evaluation used an adapted EDC analysis that measured classification error against human labels, not a true biometric FNMR from a recognition system. This approach was chosen to create a recognizer-agnostic evaluation, but future work should validate these findings with a true FNMR from one or more face recognition systems to confirm the operational benefit.

Finally, the evaluation's ground truth was established by a single rater. While this ensures perfect consistency for this study, a more generalized ground truth would require a larger-scale study involving multiple raters. Validating the metric's absolute accuracy would ideally involve data from motion-capture rigs providing known 3-D shoulder angles.

\section{Conclusion}

This paper addressed the need for a quantitative method to evaluate shoulder presentation in biometric identity documents, as mandated by emerging international standards. A Shoulder Presentation Evaluation (SPE) algorithm was proposed, operating by calculating two distinct metrics—one for horizontal alignment (yaw) and one for tilt (roll)—using only the 3D coordinates of the left and right shoulder landmarks.

The evaluation, conducted on a test set of 121 portraits, demonstrated a strong correlation ($r \approx 0.80$) between the algorithm's scores and human-assigned labels for shoulder yaw. Furthermore, an adapted Error-versus-Discard (EDC) analysis confirmed the metric's utility for compliance filtering, showing that discarding images with the lowest SPE scores effectively reduces the classification error rate against human-adjudicated labels.

This work also highlights key areas for future development. For full ISO/IEC compliance, the SPE algorithm should be integrated with an established head-pose checker. To confirm its operational impact, the adapted EDC analysis should be extended with a true FNMR evaluation using state-of-the-art face recognizers. Furthermore, a more generalized ground-truth benchmark could be established through a multi-rater study.

In conclusion, the proposed SPE algorithm is presented as a viable and effective tool for the automated assessment of shoulder presentation, filling a notable gap in current biometric quality assurance systems.

\end{document}